\documentclass[conference]{IEEEtran}
\usepackage{hyperref}
\usepackage{cite}
\usepackage{amsmath,amssymb,amsfonts}
\usepackage{algorithmic}
\usepackage{graphicx}
\usepackage{textcomp}
\usepackage{xcolor}
\usepackage{bbm}
\usepackage{booktabs}
\usepackage{caption}
\setkeys{Gin}{width=0.9\linewidth}  

\def\BibTeX{{\rm B\kern-.05em{\sc i\kern-.025em b}\kern-.08em
    T\kern-.1667em\lower.7ex\hbox{E}\kern-.125emX}}
\begin{document}

\title{Uncovering a Winning Lottery Ticket with Continuously Relaxed Bernoulli Gates}

\author{\IEEEauthorblockN{Itamar Tsayag}
\IEEEauthorblockA{\textit{Faculty of Engineering} \\
\textit{Bar-Ilan University}\\
Ramat-Gan, Israel \\
itamar.tsay@gmail.com}
\and
\IEEEauthorblockN{Ofir Lindenbaum}
\IEEEauthorblockA{\textit{Faculty of Engineering} \\
\textit{Bar-Ilan University}\\
Ramat-Gan, Israel \\
ofirlin@gmail.com}
}

\maketitle

\begin{abstract}
Over-parameterized neural networks incur prohibitive memory and computational costs for resource-constrained deployment. The Strong Lottery Ticket (SLT) hypothesis suggests that randomly initialized networks contain sparse subnetworks achieving competitive accuracy \textit{without weight training}. Existing SLT methods, notably \textsf{edge-popup}, rely on non-differentiable score-based selection, limiting optimization efficiency and scalability. We propose using continuously relaxed Bernoulli gates to discover SLTs through fully differentiable, end-to-end optimization—training only gating parameters while keeping all network weights frozen at their initialized values. Continuous relaxation enables direct gradient-based optimization of an $\ ell_0$-regularization objective, eliminating the need for non-differentiable gradient estimators or iterative pruning cycles. To our knowledge, this is the first fully differentiable approach for SLT discovery that avoids straight-through estimator approximations. Experiments across fully connected networks, CNNs (ResNet, Wide-ResNet), and Vision Transformers (ViT, Swin-T) demonstrate up to 90\% sparsity with minimal accuracy loss—nearly double the sparsity achieved by edge-popup at comparable accuracy—establishing a scalable framework for pre-training network sparsification.
\end{abstract}


\section{Introduction}
The increasing complexity and scale of modern deep learning models have significantly amplified their computational demands. While larger models generally achieve higher accuracy, their substantial resource requirements limit their training \cite{refael2025sumo,refael2025adarankgrad} and accessibility for many practitioners. 

A key factor contributing to these demands is the over-parameterization commonly observed in large neural networks \cite{rahimiunveiling}. Over-parameterized models often contain redundant components that do not substantially improve performance but still consume considerable computational resources \cite{yousefzadeh2022over}. To address this inefficiency, neural network compression techniques - such as pruning \cite{lecun1989optimal, hassibi1993optimal, zhu2017prune, gale2019state}, quantization \cite{das2018mixed}, knowledge distillation \cite{bucilua2006model, bagherinezhad2018label}, and low-rank factorization \cite{rigamonti2013learning, jaderberg2014speeding} - have been developed. These methods aim to reduce memory usage and computational costs while maintaining accuracy, making them essential for deploying models on resource-constrained devices.

Among these techniques, pruning has gained significant attention for identifying and removing unnecessary parameters while preserving the network's predictive capabilities. This aligns closely with the \textit{Lottery Ticket Hypothesis} (LTH) \cite{frankle2018lottery}, which posits that within large, over-parameterized networks, there exist smaller subnetworks - termed ``winning tickets'' - that can achieve performance comparable to the original model.

An intriguing discovery by \cite{zhou2019deconstructing} revealed that randomly selected subnetworks designated as winning tickets can exhibit non-trivial accuracy \textit{without any weight training}. Building on this, \cite{malach2020proving} demonstrated that sufficiently over-parameterized networks inherently contain subnetworks capable of achieving near-target accuracy without additional training, effectively equating pruning to training; similar principles are further leveraged in \cite{svirsky2026train} to enable efficient fine-tuning and compression via structured sparsity. This distinction gave rise to the terms \textit{weak} lottery tickets, which require further training to achieve optimal performance, and \textit{strong} lottery tickets (SLTs), which achieve competitive accuracy without modifying the original weights.

The existence of strong lottery tickets was empirically validated by \cite{ramanujan2020s}, where a subnetwork in a randomly initialized Wide-ResNet50 matched the classification accuracy of a trained ResNet34 on ImageNet. Their edge-popup algorithm remains the primary approach for identifying SLTs. However, this method has notable limitations: it requires calculating ``popup scores'' for each weight and relies on a non-differentiable gradient estimator, which makes optimization inefficient and limits scalability to larger architectures.

\textbf{Our Contribution.} We propose a fundamentally different approach to discovering strong lottery tickets. Unlike existing methods that rely on non-differentiable estimators or iterative prune-train cycles, we introduce continuously relaxed Bernoulli gates~\cite{yamada2020feature,yang2023multi,jana2023support} that enable \textit{fully 
differentiable}, end-to-end identification of sparse subnetworks with weights frozen at initialization. To our knowledge, this is the first approach to discover strong lottery tickets by continuously relaxing binary gates, eliminating the need for non-differentiable gradient estimation while achieving competitive sparsity-accuracy trade-offs across convolutional and transformer architectures.

\section{Related Work}

\subsection{Lottery Ticket Hypothesis}
The Lottery Ticket Hypothesis (LTH), introduced by \cite{frankle2018lottery}, posits that within a randomly initialized dense neural network lies a sparse subnetwork - termed a \textit{winning ticket} - that can be trained in isolation to match the accuracy of the original network. Winning tickets are typically identified through iterative magnitude-based pruning (MBP), where weights with the lowest magnitude are pruned, and the remaining weights are reset to their initial values. While effective, this process is computationally expensive due to repeated pruning and retraining cycles.

\subsection{Strong Lottery Tickets}
Building on LTH, \cite{ramanujan2020s} introduced \textit{strong lottery tickets} (SLTs) - subnetworks that achieve competitive accuracy without any weight training. Using the \textsf{edge-popup} algorithm, they demonstrated that a randomly initialized Wide ResNet-50 contains a subnetwork matching the performance of a trained ResNet-34 on ImageNet. Their work highlighted the critical role of initialization, with Scaled Kaiming Normal distribution yielding optimal results.

The theoretical foundations for SLTs were established by \cite{malach2020proving}, who proved that sufficiently overparameterized networks contain subnetworks that approximate any target network solely through pruning. Pensia et al. \cite{pensia2020optimal} later relaxed these requirements, showing that any ReLU network with width $n$ and depth $l$ can be $\epsilon$-approximated by sparsifying a random network that is $O(\log(nl))$ times wider and twice as deep.

\subsection{Neural Network Pruning}
Pruning methods can be categorized by \textit{when} they operate. \textit{Post-training} methods sparsify networks after training, including magnitude pruning \cite{han2015learningweightsconnectionsefficient}, movement pruning \cite{sanh2020movementpruningadaptivesparsity}, and structured pruning approaches. These methods typically require a trained network as input. \textit{During-training} methods integrate pruning into the training process, often using $\ell_1$ regularization to encourage sparsity. However, $\ell_1$ regularization fails to achieve true sparsity during optimization and requires post-hoc thresholding \cite{scardapane2017group}.

\textit{Pre-training} (or ``at-initialization'') methods, including our approach, identify sparse subnetworks before any weight training occurs. The edge-popup algorithm \cite{ramanujan2020s} falls into this category but relies on non-differentiable score-based selection, limiting optimization efficiency.

\subsection{Continuous Relaxations for Discrete Selection}
Gradient-based optimization over discrete variables (e.g., binary gates) requires continuous relaxations. The Concrete (or Gumbel-Softmax) distribution \cite{maddison2017concretedistributioncontinuousrelaxation, jang2017categoricalreparameterizationgumbelsoftmax} provides a differentiable approximation to discrete random variables. \cite{louizos2018learningsparseneuralnetworks} extended this with Hard-Concrete gates for $\ell_0$ regularization in network sparsification. However, these logistic-based relaxations suffer from high gradient variance due to heavy-tailed distributions.

\cite{yamada2020feature} proposed Stochastic Gates (STG) using Gaussian-based relaxations, which achieve lower variance and more stable optimization compared to logistic alternatives. Originally developed for feature selection, STG has not previously been applied to discovering strong lottery tickets. Our work bridges this gap by leveraging STG's favorable optimization properties to pre-train neural networks for sparsification.

\section{Proposed Solution}
We present a novel approach for uncovering strong lottery tickets (SLTs) by leveraging continuously relaxed Bernoulli variables as differentiable gating mechanisms \cite{yamada2020feature}. These gates enable unstructured weight pruning while maintaining end-to-end differentiability. Importantly, the original network weights $W^{(i)}$ remain frozen throughout training; only the gating parameters are optimized, similar in spirit to Edge-Popup \cite{ramanujan2020s}. The gating network selectively identifies important weights by applying learnable gating variables at the weight level, facilitating efficient sparsification while preserving model expressiveness.

The gating variable for a weight between neurons $i$ and $j$ in layer $l$ is defined as:
\[
z_{ij}^l = \max(0, \min(1, \mu_{ij}^l + \epsilon_{ij}^l)),
\]
where $\epsilon_{ij}^l \sim \mathcal{N}(0, \sigma_{CRBG}^2)$ represents Gaussian noise, $\mu_{ij}^l$ is a learned parameter, and the hard-sigmoid function ensures $z_{ij}^l \in [0, 1]$.

The stochastic component $\epsilon$ serves a critical purpose: it enables gradient-based optimization despite the discrete nature of gates via continuous relaxation, as established in prior work \cite{yamada2020feature, louizos2018learningsparseneuralnetworks} and used successfully in many applications \cite{naorhybrid,lindenbaum2021l0}. Unlike $\ell_1$ regularization, which fails to achieve true sparsity during training and requires post-hoc thresholding \cite{scardapane2017group}, our approach achieves exact zeros. The resampled noise also prevents premature pruning by allowing gates to reactivate during optimization.

In unstructured pruning, weight-level gates $B^{(i)}$ selectively mask individual weights in layer $i$:
\[
\tilde{G}^{(i)}(h^{(i-1)}) = \sigma(B^{(i)} \odot W^{(i)} h^{(i-1)}),
\]
where $B^{(i)}$ is a matrix of gating variables matching the dimensions of $W^{(i)}$, $h^{(i-1)}$ denotes the input to layer $i$ (i.e., the output of the previous layer), $\sigma$ is the activation function, and $\odot$ denotes element-wise multiplication.

The overall network $G(x)$ is defined as:
\[
G(x) = \tilde{G}^{(L)} \circ \cdots \circ \tilde{G}^{(1)}(x),
\]
where $L$ is the number of layers.

To induce sparsity, the objective function incorporates regularization on the gating parameters only (weights remain frozen):
\[
\min_{\{B^{(i)}\}} \mathcal{L} \left( \{ B^{(i)} \odot W^{(i)} h^{(i-1)} \}_{i=1}^L \right) + \lambda \sum_{i=1}^L \mathbb{E}[\| B^{(i)} \|_0].
\]
By taking the expectation, the non-differentiable $\ell_0$ term becomes differentiable:
\[
\mathbb{E}[\| B^{(i)} \|_0] = \sum_{j,k} P(B_{jk}^{(i)} \neq 0) = \sum_{j,k} \Phi\left(\frac{\mu_{jk}^{(i)}}{\sigma_{CRBG}}\right),
\]
where $\Phi$ is the standard Gaussian CDF. This formulation enables gradient-based optimization while directly penalizing the expected number of active gates.

This method provides a differentiable, efficient pruning strategy that achieves exact sparsity without post-hoc thresholding, supporting unstructured sparsification for flexible network compression.

\paragraph{Inference.} After training, stochastic noise is removed by setting $\epsilon = 0$. The final binary mask is obtained by thresholding: $\hat{z}_{jk}^{(i)} = \mathbbm{1}[\mu_{jk}^{(i)} > 0]$, where $\mathbbm{1}[\cdot]$ is the indicator function. Weights with $\hat{z} = 0$ are pruned, yielding a sparse, deterministic subnetwork for deployment.

\section{Experiments}

This section evaluates our method for identifying Strong Lottery Tickets (SLTs) using continuously relaxed Bernoulli gates. All experiments focus on \textit{pre-training sparsification}: base network weights are randomly initialized and remain frozen throughout; only the gating parameters are optimized. Experiments span fully connected networks (FCNs), convolutional neural networks (CNNs), and Transformer architectures.

\subsection{Experimental Setup}

\paragraph{Hyperparameters.} Following \cite{yamada2020feature}, we fix $\sigma = 0.5$ across all experiments. The regularization strength $\lambda$ was selected via grid search on a validation set, yielding $\lambda = 0.1$ for LeNet-300-100, $\lambda = 0.05$ for ResNet50, and $\lambda = 0.08$ for Transformers. These values were then held fixed for all reported results.

\paragraph{Training Details.} Gating parameters $\mu$ are initialized to $0.5$ and optimized using Adam with learning rate $10^{-3}$. Training runs for 100 epochs (FCNs), 50 epochs (CNNs), and 30 epochs (Transformers). All experiments use a single NVIDIA RTX 3090 GPU.

\paragraph{Evaluation Protocol.} We report top-1 accuracy and sparsity (percentage of pruned weights). For fair comparison, we distinguish between methods that \textit{train weights} (weak LT methods) and those that \textit{freeze weights} (strong LT methods, including ours).

\subsection{Sparsification of Fully Connected Networks}

We evaluate on LeNet-300-100 \cite{lecun1998gradient} using the MNIST dataset. A gating network with the same architecture as the base network applies element-wise masks to the frozen base weights. The $\ell_0$ regularization term (Section 3) encourages sparsification.

\begin{figure}[!ht]
    \centering
    \includegraphics[]{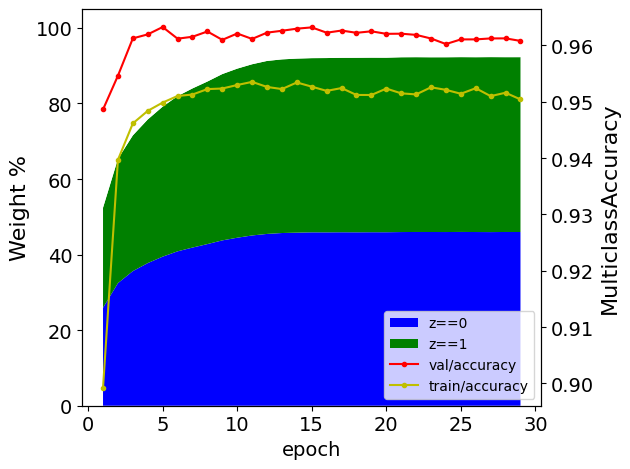}
    \vspace{-5pt}
    \caption{\textbf{Pre-training sparsification on LeNet-300-100.} Blue region: percentage of pruned weights. Green region: retained weights. Red line: test accuracy on MNIST as sparsification progresses. The method achieves 96\% accuracy at 45\% sparsification.}
    \label{fig:strong_lt_sparsification}
\end{figure}

As shown in Fig.~\ref{fig:strong_lt_sparsification}, our method achieves \textbf{96\% accuracy with 45\% sparsification}, aligning with findings from \cite{ramanujan2020s} that approximately 50\% sparsification provides an optimal accuracy-sparsity trade-off for SLTs.

Table~\ref{ComparativeResultsFC} compares our method against prior work. Methods above the dashed line \textit{train weights} and thus identify weak lottery tickets; methods below identify \textit{strong} lottery tickets without weight modification. Our method significantly outperforms the edge-popup variant from \cite{xiong2022stronglotterytickethypothesis}, which achieved only 85\% accuracy at 50\% sparsification on a larger base network (500-500-500-500).

\begin{figure}[!ht]
    \centering
    \includegraphics[]{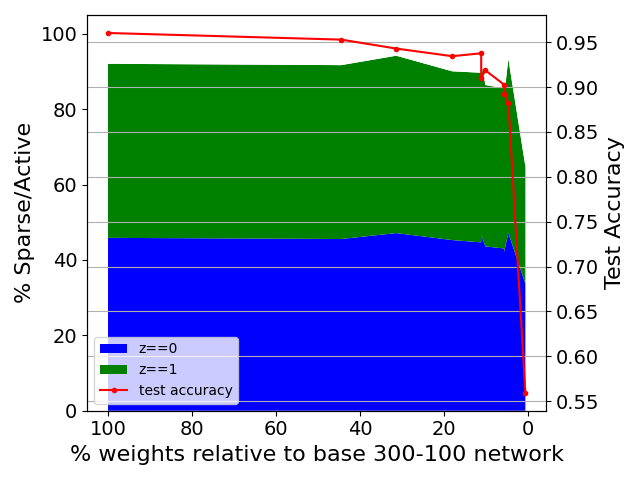}
    \vspace{-5pt}
    \caption{Robustness to base network size (LeNet on MNIST). Blue: pruned weights. Green: retained weights. Red: test accuracy. SLTs can be discovered even in base networks at 20\% of the original size.}
    \label{fig:StrongLTBaseNetwrkSize}
\end{figure}

Fig.~\ref{fig:StrongLTBaseNetwrkSize} demonstrates robustness: SLTs can be identified even in base networks reduced to 20\% of the original size, maintaining competitive accuracy.

\begin{table*}[htp]
    \centering
    \renewcommand{\arraystretch}{1.2}
    \resizebox{\textwidth}{!}{ 
    \begin{tabular}{@{}llcccc@{}} 
        \toprule
        \textbf{Method} & \textbf{Description} & \textbf{Architecture} & \textbf{Accuracy} & \textbf{Weights Trained} & \textbf{Sparsity} \\ 
        \midrule
        \multicolumn{6}{l}{\textit{Weak Lottery Ticket Methods (weights trained)}} \\
        Baseline \cite{lecun1998gradient} & Standard training with $\ell_2$ loss & 784-300-100 & 96.95\% & Yes & 0\% \\
        Sparse VD \cite{molchanov2017variational} & Variational dropout sparsification & 512-114-72 & 98.2\% & Yes & 97.8\% \\
        BC-GNJ \cite{louizos2017bayesian} & Bayesian compression with GNJ prior & 278-98-13 & 98.2\% & Yes & 89.2\% \\
        BC-GHS \cite{louizos2018learningsparseneuralnetworks} & Bayesian compression with GHS prior & 311-86-14 & 98.2\% & Yes & -- \\
        \midrule
        \multicolumn{6}{l}{\textit{Strong Lottery Ticket Methods (weights frozen)}} \\
        Edge-popup \cite{xiong2022stronglotterytickethypothesis} & Score-based selection + $\epsilon$-perturbation & 500-500-500-500 & 85\% & No & 50\% \\ 
        \textbf{Ours (CRBG)} & Continuously relaxed Bernoulli gates & 784-300-100 & \textbf{96\%} & No & 45\% \\
        \bottomrule
    \end{tabular}}
    \vspace{-7pt}
    \caption{Comparison of pruning methods on MNIST. Methods are grouped by whether weights are trained (weak LT) or frozen (strong LT). Our method achieves the highest accuracy among SLT methods while using a smaller base network than \cite{xiong2022stronglotterytickethypothesis}.}
    \label{ComparativeResultsFC}
\end{table*}

\subsection{Sparsification of CNNs}

We evaluate on CIFAR-10 using ResNet50 and Wide-ResNet50 architectures. Base networks are initialized with random weights (Scaled Kaiming Normal \cite{ramanujan2020s}) and remain frozen.

\begin{figure}[!ht]
    \centering
    \includegraphics[]{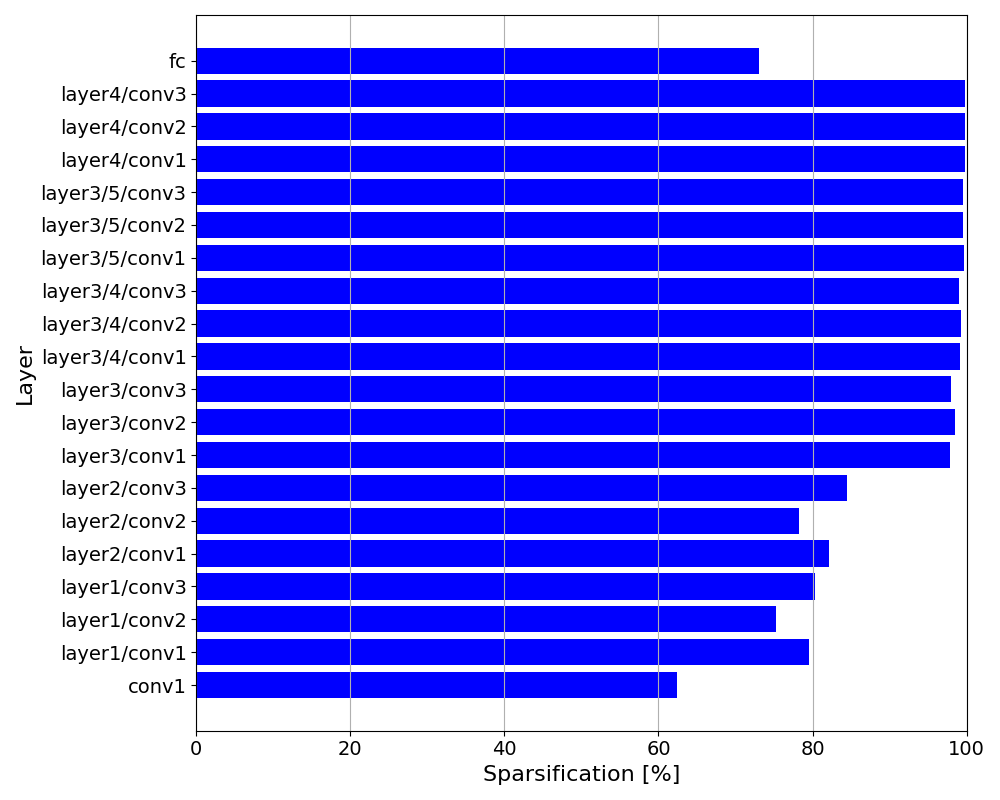}
    \vspace{-5pt}
    \caption{Per-layer sparsification of ResNet50 on CIFAR-10. Later layers exhibit higher sparsification rates, consistent with prior findings that early layers require more weights for low-level feature extraction.}
    \label{fig:resnet50_layer_sparsification}
\end{figure}

\paragraph{ResNet50.} Our method achieves \textbf{83.1\% top-1 accuracy with 91.5\% sparsification}. Fig.~\ref{fig:resnet50_layer_sparsification} shows per-layer sparsification rates, revealing higher sparsification in later layers - consistent with observations in \cite{sreenivasan2022raregemsfindinglottery} that earlier layers retain more weights due to their role in low-level feature extraction.

\paragraph{Wide-ResNet50.} Achieves \textbf{88\% top-1 accuracy with 90.5\% sparsification}, demonstrating that wider architectures provide a richer search space for discovering SLTs.

\paragraph{Comparison with Edge-Popup.} Table~\ref{tab:cnn_comparison} compares against the original edge-popup algorithm \cite{ramanujan2020s}. While edge-popup achieves similar accuracy, it requires significantly lower sparsification (50\% vs. our 90\%+). Our method achieves nearly double the sparsification at comparable accuracy, demonstrating the effectiveness of continuous relaxation over non-differentiable score-based selection.

\begin{table}[ht]
    \centering
    \begin{tabular}{@{}lccc@{}}
        \toprule
        \textbf{Method} & \textbf{Architecture} & \textbf{Accuracy} & \textbf{Sparsity} \\
        \midrule
        Edge-popup \cite{ramanujan2020s} & Wide-ResNet50 & 88\% & 50\% \\
        \textbf{Ours (CRBG)} & Wide-ResNet50 & 88\% & \textbf{90.5\%} \\
        \midrule
        \textbf{Ours (CRBG)} & ResNet50 & 83.1\% & 91.5\% \\
        \bottomrule
    \end{tabular}
    \caption{CNN sparsification on CIFAR-10. Our method matches edge-popup accuracy while achieving significantly higher sparsification.}
    \label{tab:cnn_comparison}
\end{table}

\subsection{Sparsification of Transformer-based Networks}

We extend our evaluation to Vision Transformers (ViT-base) and Swin Transformers (Swin-T) on CIFAR-10, demonstrating applicability to attention-based architectures.

\paragraph{ViT-base.} Achieves \textbf{76\% top-1 accuracy with 90\% sparsification}. To our knowledge, no prior work has explicitly targeted strong lottery tickets in Vision Transformers; existing studies focus on weak lottery tickets that require weight training \cite{shen2023datalevellotteryticket} or on post-training pruning followed by fine-tuning \cite{prasetyo2023sparse}.

\paragraph{Swin-T.} Achieves \textbf{80\% top-1 accuracy with 50\% sparsification}. For reference, a fully trained Swin-T achieves 87.26\% on CIFAR-10 - our SLT retains 92\% of full model performance without any weight training.

\paragraph{Comparison with Existing Methods.} Table~\ref{tab:transformer_comparison} contextualizes our results. While direct comparison is limited (no prior SLT methods exist for Transformers), we outperform weak LT methods at comparable sparsity levels, despite not training any weights.

\begin{table}[ht]
    \centering
    \resizebox{\linewidth}{!}{%
    \begin{tabular}{@{}lcccc@{}}
        \toprule
        \textbf{Method} & \textbf{Architecture} & \textbf{Accuracy} & \textbf{Sparsity} & \textbf{Weights Trained} \\
        \midrule
        Fully trained & Swin-T & 87.26\% & 0\% & Yes \\
        \cite{prasetyo2023sparse} & ViT-base & 85.09\% & 30\% & Yes (after pruning) \\
        \midrule
        \textbf{Ours (CRBG)} & ViT-base & 76\% & 90\% & No \\
        \textbf{Ours (CRBG)} & Swin-T & 80\% & 50\% & No \\
        \bottomrule
    \end{tabular}%
    }
    \caption{Transformer sparsification on CIFAR-10.}
    \label{tab:transformer_comparison}
\end{table}

\subsection{Summary}

Our experiments demonstrate that continuously relaxed Bernoulli gates effectively discover strong lottery tickets across diverse architectures - FCNs, CNNs, and Transformers - achieving high sparsity without weight training. Key findings:
\begin{itemize}
    \item On FCNs, we outperform prior SLT methods by 11\% accuracy (96\% vs. 85\%).
    \item On CNNs, we achieve 90\%+ sparsification compared to 50\% for edge-popup at similar accuracy.
    \item On Transformers, we establish the first SLT results, retaining 87--92\% of full model performance.
\end{itemize}

\section{Conclusion and Future Work}

This work explored the potential of uncovering strong lottery ticket subnetworks using continuously relaxed Bernoulli gates. We demonstrated that, within randomly initialized, overparameterized neural networks, it is possible to identify subnetworks that achieve competitive performance with fully trained networks. Our findings highlighted the capability of a gating mechanism based on relaxed Bernoulli variables to achieve high sparsity levels with minimal loss in accuracy, resulting in significant reductions in model size and computational demands.

The experiments with LeNet, ResNet50, and ViT-base validated the effectiveness of our pruning method in unstructured settings, underscoring its promise as a neural network compression approach. By leveraging a gating mechanism based on relaxed Bernoulli variables, we can identify strong lottery-ticket subnetworks without extensive fine-tuning, laying the groundwork for efficient pruning methods that complement traditional training.

Future research could aim to refine our framework by optimizing gating parameters or incorporating adaptive mechanisms with trainable hyperparameters to dynamically balance sparsity and accuracy. Additionally, extending our pruning method to novel architectures, such as graph neural networks and recurrent networks, and exploring applications to multi-modal settings \cite{lindenbaum2015learning}.

Additionally, exploring more sophisticated gating mechanisms could improve pruning precision and flexibility. Multi-level gating strategies, where gates produce outputs beyond binary values, could allow for finer control over network sparsity, preserving expressiveness while optimizing performance. Such advancements could refine the trade-off between sparsity and accuracy, enabling more adaptive and efficient pruning techniques.

In summary, this research demonstrates that using relaxed Bernoulli variables as a gating network offers a robust and efficient method for neural network pruning, unlocking the potential of lottery ticket subnetworks and paving the way for further innovation in network compression and optimization.

\bibliographystyle{IEEEtran}
\bibliography{references}

\end{document}